\documentclass[conference,a4paper]{IEEEtran}
\IEEEoverridecommandlockouts

\usepackage[hidelinks]{hyperref}
\usepackage[cmex10]{amsmath}
\usepackage{amssymb,amsfonts}
\usepackage{dblfloatfix}

\usepackage[ruled,vlined]{algorithm2e}
\usepackage{graphicx}
\graphicspath{{Figures/PDF/}{Figures/PNG/}}

\usepackage{booktabs}
\usepackage{siunitx}
\usepackage[numbers,compress]{natbib}
\usepackage{texnames}
\usepackage{bm,bbm}
\usepackage{orcidlink}

\usepackage{pifont}

\begin{document}

\title{\uppercase{CSFMamba: Cross State Fusion Mamba Operator for Multimodal Remote Sensing Image Classification}
}

\author{	\IEEEauthorblockN{Qingyu Wang\orcidlink{0009-0007-8869-6901}}
	\IEEEauthorblockA{\textit{Shanghai Jiao Tong University}\\
		200240 Shanghai, China\\
		wqy616@sjtu.edu.cn}
	\and
	\IEEEauthorblockN{Xue Jiang\orcidlink{0000-0001-7099-6817}}        
	\IEEEauthorblockA{\textit{Shanghai Jiao Tong University}\\
		200240 Shanghai, China\\
		xuejiang@sjtu.edu.cn}
	\and
	\IEEEauthorblockN{Guozheng Xu\orcidlink{0000-0002-5083-5896}}       
	\IEEEauthorblockA{\textit{Shanghai Jiao Tong University}\\
		200240 Shanghai, China\\
		guozhengxu@sjtu.edu.cn}
}

\maketitle
\begin{abstract}
	Multimodal fusion has made great progress in the field of remote sensing image classification due to its ability to exploit the complementary spatial-spectral information. Deep learning methods such as CNN and Transformer have been widely used in these domains. State Space Models recently highlighted that prior methods suffer from quadratic computational complexity. As a result, modeling longer-range dependencies of spatial-spectral features imposes an overwhelming burden on the network. Mamba solves this problem by incorporating time-varying parameters into ordinary SSM and performing hardware optimization, but it cannot perform feature fusion directly. In order to make full use of Mamba's low computational burden and explore the potential of internal structure in multimodal feature fusion, we propose Cross State Fusion Mamba (CSFMamba) Network. Specifically, we first design the preprocessing module of remote sensing image information for the needs of Mamba structure, and combine it with CNN to extract multi-layer features. Secondly, a cross-state module based on Mamba operator is creatively designed to fully fuse the feature of the two modalities. The advantages of Mamba and CNN are combined by designing a more powerful backbone. We capture the fusion relationship between HSI and LiDAR modalities with stronger full-image understanding. The experimental results on two datasets of MUUFL and Houston2018 show that the proposed method outperforms the experimental results of Transformer under the premise of reducing the network training burden.
\end{abstract}

\begin{IEEEkeywords}
	Convolutional neural networks, hyperspectral/LiDAR (HSI/LiDAR) data, State Space Model (SSM), data preprocessing, cross-state.
\end{IEEEkeywords}

\section{Introduction}

Remote sensing (RS) imaging is a sensor detection technology for target objects from a long distance, which has many applications in land cover classification and other fields \cite{yue2022Spectral, jiang2022afusion} . Diverse data fusion, such as between hyperspectral imagery (HSI) and light detection and Ranging (LiDAR) can form spatial-spectral feature groups. This can obtain a more adequate description of the surface information and greatly improve the accuracy of RS image classification \cite{xu2023robust, xu2023exploring}.

\begin{figure*}[hbt]
	\centering
	\includegraphics[width=\linewidth]{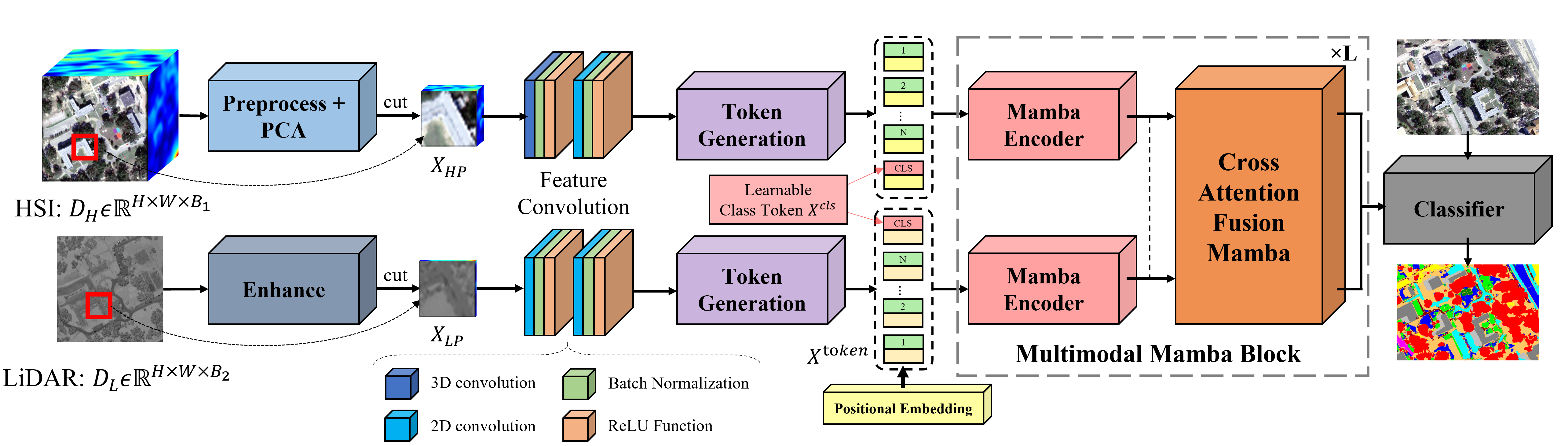}
	\caption{Illustration of the overall structure of our proposed CSFMamba.}\label{NetworkStructure}
\end{figure*}

In the earliest studies of multimodal fusion, Support Vector Machine (SVM) \cite{melgani2004classification} and Random Forest (RF) \cite{ahmad2021hyper} can extract the most basic features. Deep learning methods, such as Convolutional Neural Networks (CNN), have gradually become mainstream methods due to their powerful feature extraction capabilities \cite{hong2020deep, hang2020classification, zhao2021fractional}. As remote sensing images become more detailed, the demand for long-term memory of the network gradually increases. The network feature extraction operator has been developed from Recurrent Neural Network (RNN) \cite{wu2017convolutional} to Transformer method \cite{vaswani2017attention}. Zhao et al. \cite{zhao2022joint} make full use of the internal structure of Transformer to fully combine attention and modality fusion. Li et al. \cite{li2023mixing} integrates self-attention and convolution to achieve better results.

However, driven by self-attention mechanism, Transformer's computation has quadratic computational complexity. This also puts a certain limit on its long-term memory ability. It is worth noting that a powerful tool has recently emerged in the field of large models, the Mamba operator \cite{gu2023mamba}. Based on the State Space Model (SSM) \cite{gu2021eff}, the attention scales linearly with the sequence length by introducing the hidden space. Yang et al. \cite{yang2024hsimamba} and Huang et al. \cite{huang2024spectral} explored the application of Mamba to object classification in remote sensing HSI images. Gao et al. \cite{gao2024multi} first introduced Mamba into multimodal remote sensing image classification and achieved certain results. However, these algorithms are difficult to make up for the heterogeneity between multimodal data, and do not extend the hidden space of Mamba to the cross field. Or simply replace the original structure with the Mamba operator, failing to achieve better results through more rational overall design.

Therefore, to address the lack of a similar cross-attention design in Mamba, this paper proposes an approach based on a cross-state fusion module. Firstly, Mamba operator has good scalability, but it lacks a good backbone of visual features. To this end, we enhance the original data of HSI and LiDAR, and combine CNN to extract local features, so as to give full play to the advantages of excellent long-term memory of the SSM. Secondly, Mamba operator suffers from feature redundancy in multiple scan paths. After we optimize the structure of Mamba operator, we further extend its advantage of reducing the computational burden. Finally, aiming at the feature differences between multi-source data, we creatively propose a cross-state module using Mamba’s time-varying parameters. By intercrossing internal state spaces, the issue of Mamba's ineffective application in high-resolution and fine-grained multimodal tasks is resolved.

\section{Methodology}

\subsection{Preprocessing and feature extraction}
We propose a Mamba-based signal feature extraction and state fusion network, which is suitable for the input of HSI and LiDAR signals, as shown in Fig.~\ref{NetworkStructure}. The Principal Component Analysis (PCA) method is designed in the HSI branch to reduce the spectral dimension. Before principal component analysis, mutual information is introduced to select the relatively important and more informative components.The LiDAR branch can analyze the curvature, gradient, local mean and variance from the height information. Through this process, the signals of the two modals are processed into $X_H \in \mathbb{R}^{H \times W \times C_1}$ and $X_L \in \mathbb{R}^{H \times W \times C_2}$.

For $H\times W$ pixels to be classified, the class feature of the pixel is obtained by analyzing the neighborhood. The network splits it into sets of small cube patches $X_{HP} \in \mathbb{R}^{s \times s \times C_1}$ and $X_{LP} \in \mathbb{R}^{s \times s \times C_2}$, where $s \times s$ is the size of each patch. Pixels on the edge will fill the surrounding area with zeros. Then it is divided into training set and validation set.

Given the strong local feature extraction capability of convolutional layers, a two-layer convolutional network is designed to extract patch features $X_{HF} \in \mathbb{R}^{s \times s \times D}$. $D$ is the output dimension of the convolution. HSI has a larger number of channels, so we designed a 3D plus 2D convolution module for it. Correspondingly, two 2D convolutions are designed for LiDAR branches with fewer channels and keep the output $X_{LF}$ the same size. Each convolutional layer is followed by a batch normalization and a ReLU activation layer.

\begin{figure*}[hbt]
	\centering
	\includegraphics[width=\linewidth]{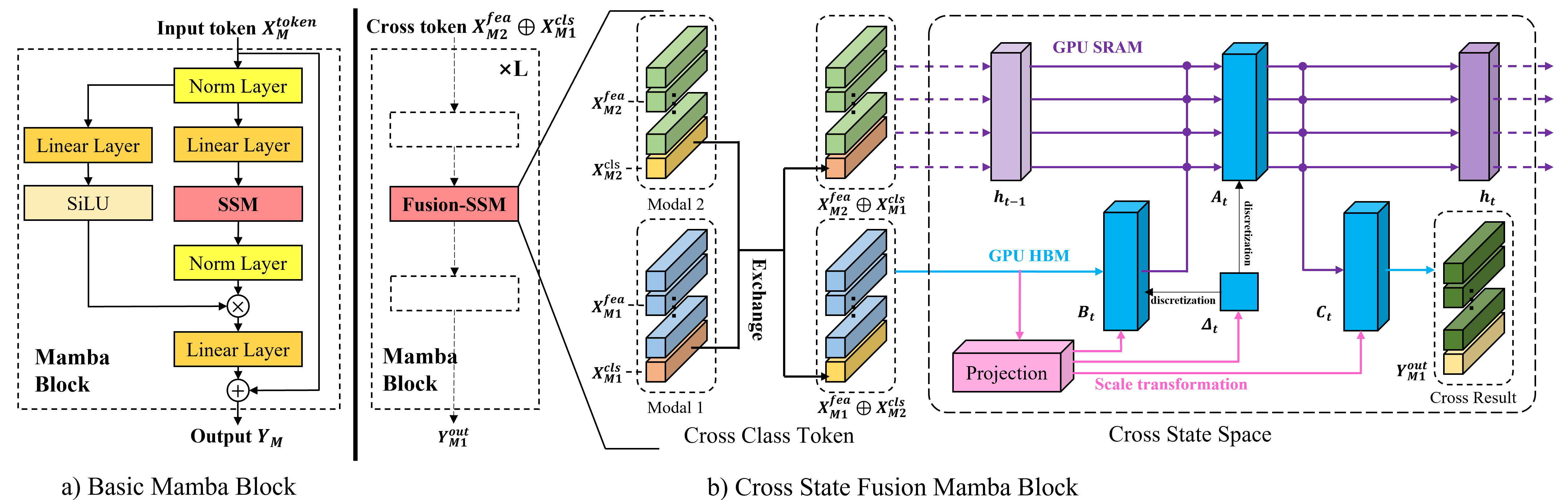}
	\caption{Framework of the proposed Mamba block and Cross State Fusion Mamba block. (a) Details of the proposed basic Mamba block simplified in common architecture. (b) Details of the proposed CSFM block. Innovations are made primarily at the SSM core of the Mamba infrastructure.}\label{CrossState}
\end{figure*}

\subsection{Feature tokenization}

The multi-scale feature obtained by the convolutional layer also needs to be further embedded to adapt to the input requirements of the Mamba module. Specifically, tokenization is a projection process. Firstly, the features $X_{HF} \in \mathbb{R}^{s \times s \times D}$ is projected by a 2D convolution, and then flattened into a set of planar feature maps. Finally, we use a linear layer to project the planarized feature map to the required dimension, obtaining the spatial feature tokens $X_H^{fea}\in\mathbb{R}^{B\times L\times D}$:
\begin{equation}
    X_H^{fea}=Flatten(X_{HF}\cdot W_{conv})\cdot W_{norm}
\end{equation}

 Where $W_{conv}$ and $W_{norm}$ represent the learnable parameters of 2D convolutional and linear layers. $B$ is the output dimension of the linear layer and $L$ is the square of the parameters of the 2D convolution kernel. After that, an all-zero learnable class token is embedded after all feature tokens. CLS Token will learn the overall abstract features of the tokens and work in the final classification. The feature token is embedded via a 2D sinusoidal position $P_H\in\mathbb{R}^{B\times(L+1)\times D}$ to supplement the position information.
\begin{equation}
    X_H^{token}=Concat(X_H^{fea},X_H^{cls})+P_H
\end{equation}

Where $X_H^{cls}\in\mathbb{R}^{B\times 1\times D}$ represents the CLS Token. Through this process, the final overall token set $X_H^{token}\in\mathbb{R}^{B\times(L+1)\times D}$ is obtained. The LiDAR branch is processed in exactly the same way, yielding $X_L^{token}\in\mathbb{R}^{B\times(L+1)\times D}$.

\subsection{Mamba encoder}
State Space Model (SSM) reduces the complexity of the operator by linear operation. Different from other algorithms, SSM utilizes the hidden state $h(t)\in \mathbb{R}^{N}$ as an intermediate variable. An ordinary differential equation is used to link the input sequence $x(t)$ and the output sequence $y(t)$.

The input-output sequence is linked by four key matrices. $\mathbf{A}\in \mathbb{R}^{N \times N}$ represents the state transition matrix. $\mathbf{B}\in \mathbb{R}^{N \times 1}$ and $\mathbf{C}\in \mathbb{R}^{N \times 1}$ are the system parameters. $\mathbf{D}\in \mathbb{R}^{N \times 1}$ as the optional residual connection matrix. $N$ denotes the number of states. In order to be used in the generation-by-generation operation of deep learning, the time parameter $\Delta$ is introduced to discretize the above matrix:
\begin{gather}
    \overline{\mathbf{A}}=\exp(\Delta\mathbf{A})\\
    \overline{\mathbf{B}}=(\Delta\mathbf{A})^{-1}(\exp(\Delta\mathbf{A})-\mathrm{I})\cdot\Delta\mathbf{B}
\end{gather}

Traditional SSM is a linear time-invariant system where the network parameter matrix is kept fixed. Mamba considers that the $\mathbf{B}$,$\mathbf{C}$,$\Delta$ matrix can be dynamically modeled through the input sequence according to the context. The parameter matrix can depend on the input, allowing individual elements of the sequence to receive more attention. Mamba's improved calculation method is shown below, and the values of the matrix are related to the input variables:
\begin{equation}
    \begin{aligned}h_t&=\overline{\mathbf{A}}\cdot h_{t-1}+\overline{\mathbf{B}}(x_t)\cdot x_t \\
    y_t&=\mathbf{C}(x_t)\cdot h_t+\mathbf{D}\cdot x_t\end{aligned}
\end{equation}

Based on the basic modules of Mamba applied to CV tasks, CSFMamba simplifies the structure of Mamba by removing Conv1D and a projection function. The structure of the Mamba feature extraction operator is shown in Fig.~\ref{CrossState}. In order to extract deep features, multi-layer Mamba is used here, where each layer has the same size of input and output.

\begin{table*}[hbt]
	\centering
	\caption{Classification performance of \textbf{MUUFL} dataset on different methods. The best is shown in bold.}\label{MUUFL result}
        \resizebox{0.8\linewidth}{!}{
	\begin{tabular}{c||cccccc}
		\toprule
		Indicators & EndNet\cite{hong2020deep} & CoupledCNN\cite{hang2020classification} & FGCN\cite{zhao2021fractional} & HCT\cite{zhao2022joint} & MACN\cite{li2023mixing} & Ours     \\ 
        \hline
        \rule{0pt}{2.2ex}
		OA(\%) & 82.84 & 87.22 & 89.04 & 88.47 & 90.66 & \textbf{91.13} \\
		AA(\%) & 85.94 & 89.14 & 90.76 & 90.89 & 92.21 & \textbf{92.52} \\
		Kappa  & 78.72 & 84.52 & 86.13 & 85.19 & 87.73 & \textbf{88.30} \\
		\bottomrule
	\end{tabular}
 }
\end{table*}

\begin{table*}[hbt]
	\centering
	\caption{Classification performance of \textbf{Houston2018} dataset on different methods. The best is shown in bold.}\label{Houston2018 result}
        \resizebox{0.8\linewidth}{!}{
	\begin{tabular}{c||cccccc}
		\toprule
		Indicators & CoupledCNN\cite{hang2020classification} & FGCN\cite{zhao2021fractional} & HCT\cite{zhao2022joint} & MACN\cite{li2023mixing} & MSFMamba\cite{gao2024multi} & Ours     \\ 
        \hline
        \rule{0pt}{2.2ex}
		OA(\%) & 75.19 & 68.40 & 82.31 & 84.58 & 92.38 & \textbf{93.38} \\
		AA(\%) & 81.32 & 66.04 & 86.86 & 87.65 & 95.51 & \textbf{96.40} \\
		Kappa  & 68.43 & 58.34 & 72.77 & 78.41 & 90.16 & \textbf{91.45} \\
		\bottomrule
	\end{tabular}
 }
\end{table*}

\begin{table}[hbt]
	\centering
	\caption{Burden analysis of model parameters and FLOPs on the MUUFL dataset. The best is shown in bold.}\label{Burden}
        \resizebox{\linewidth}{!}{
	\begin{tabular}{c||cccccc}
		\toprule
		Indicators & CoupledCNN & FGCN & HCT & MACN & Ours     \\ 
        \hline
        \rule{0pt}{2.2ex}
		params(M) & 0.450 & 1.001 & 0.582 & 0.318 & \textbf{0.219} \\
        FLOPs(M)  & 15.32 & 19.60 & 16.09 & 18.11 & \textbf{12.29} \\
		\bottomrule
	\end{tabular}
 }
\end{table}

\begin{table}[hbt]
	\centering
	\caption{Ablation study on main innovation module with results highlighting the $OA$ metric. The best is shown in bold.}\label{Ablation}
        \resizebox{0.9\linewidth}{!}{
	\begin{tabular}{c|ccc|c}
		\toprule
		Cases & Preprocess & Conv2/3D & Cross-State & OA(\%)     \\ 
        \hline
        \rule{0pt}{2.2ex}
		1 & \ding{55} & \ding{52} & \ding{52} & 89.68          \\
		2 & \ding{52} & \ding{55} & \ding{52} & 88.61          \\
        3 & \ding{52} & \ding{52} & \ding{55} & 82.94          \\
		4 & \ding{52} & \ding{52} & \ding{52} & \textbf{91.13} \\
		\bottomrule
	\end{tabular}
        }
\end{table}

\subsection{Cross State Fusion Mamba block}
Feature fusion is the most important part of multimodal RS image classification. Therefore, we creatively design the Cross State Fusion Mamba block to fuse the deep features of HSI and LiDAR. Specifically, we utilize CLS Token as a proxy for modal features to exchange information with another modality. The CLS Token has fully learned all the abstract information of a modality in the Mamba Encoder of the previous step. We take the CLS Token from one modality and splice it into the corresponding position of the other modality:
\begin{equation}
    X_H^{cross}=X_L^{fea}\cup X_H^{cls}
\end{equation}

The newly obtained overall Token will be fed into the Mamba block again to extract features, so that the mutual information of the two modalities can be fully utilized. We note that the parameter generation process of SSM in Mamba can also form a crossover process to achieve the crossover of states. As introduced in Section II.C, the parameter matrix can also be informed from another modality:
\begin{equation}
    \begin{aligned}
    h_t&=\overline{\mathbf{A}}\cdot h_{t-1}+\overline{\mathbf{B}}(X^{cross}_{L,t})\cdot X^{cross}_{H,t} \\
    Y_{H,t}&=\mathbf{C}(X^{cross}_{L,t})\cdot h_t+\mathbf{D}\cdot X^{cross}_{H,t}
    \end{aligned}
\end{equation}

The overall process of Cross State Fusion Mamba block is shown in Fig.~\ref{CrossState}. Combined token set goes through the basic Mamba block to extract information, but the new state is formed using cross input in the fusion-SSM. HSI branch refers to the set of tokens that have $X_H^{cls}$, which is fused with the LiDAR modal state space to obtain the final feature $Y_H^{out}$:
\begin{equation}
    Y_H^{out}=Norm(CSFMamba(X_H^{cross},X_L^{cross}))
\end{equation}

The complexity is greatly reduced due to the whole process linearity. CSFMamba uses an $L$ multi-layer structure. LiDAR branch can obtain the results $Y_L^{out}$ with  the same operation.

\subsection{Classification block}
At the end of the network, CLS tokens of two modalities are extracted to participate in classification. $Y_H^{cls}$and $Y_L^{cls}$are averaged and mapped to the label probability space through a fully connected layer. The output $\hat{p}\in\mathbb{R}^K=FC((Y_H^{cls}+Y_L^{cls})/2)$ represents the probability of each pixel belonging to various categories. $K$ is the number of classes in the dataset. Finally, the model uses cross-entropy loss to optimize the model:
\begin{equation}
    \mathcal{L}_{cls}=-\sum_{k=1}^{K}p_{k}\log \hat{p}_{k}
\end{equation}

where $\hat{p}_{k}$ is the k-th element of $\hat{p}$, and $p_{k}$ is the k-th element of the one-hot label vector. Through the overall process as shown in Fig.~\ref{NetworkStructure}, the classification of each pixel in the whole picture is realized.

\section{Results}

\subsection{Experimental Setup}
In this experiment, two relatively large datasets of multimodal RS image target classification are selected to prove the adaptability of the network in complex tasks. The picture size of MUUFL dataset is 325×220 pixels, with 72 bands of HSI data and 11 categories. The image of Houston2018 dataset consists of 1202×4768 pixels and has 20 categories. This experiment is carried out on a RTX3090-24GB GPU. The overall network is built on python3.11 and pytorch framework. Batch size is 256 using Adam optimizer. We set the initial learning rate to 5e-4 and the decay rate to 0.5. The patch size is taken to be 11×11, and the training process has 200 iterations in total.

\subsection{Main Results}

We evaluate the performance of the proposed Mamba operator on two complex datasets, MUUFL and Houston2018. The methods based on CNN, Transformer and direct Mamba operator are taken into consideration for comparison. We measure the performance of the network with three general image classification metrics: overall accuracy ($OA$), average accuracy ($AA$), and kappa coefficient ($\kappa$).

CSFMamba achieves competitive results on multimodal image classification with HSI and LiDAR source information. On the MUUFL dataset, CSFMamba achieves the highest $OA$ of 91.13\%, $AA$ of 92.52\%, and $\kappa$ of 88.30\%, as in Table~\ref{MUUFL result}. It performs better on Houston2018 dataset, and obtains the highest $OA$ of 93.38\%, $AA$ of 96.40\%, and $\kappa$ of 91.45\%, as in Table~\ref{Houston2018 result}. Our network achieves remarkable results and has outstanding capabilities in tasks involving complex scenarios.

Moreover, using Mamba operator imposes less burden on the network. The experiment Table~\ref{Burden}. shows the lightweight and low burden of the network by comparing the number of parameters and Floating Point Operations per Second (FLOPs). The linear computation of SSM theoretically saves a lot of memory and computational space. The amount of computation of CSFMamba is 12.29MFLOPs, and the number of parameters is 0.219M, which is much smaller than CNN-based and transformer-based methods. At the same time, the training and testing time is shorter. This method achieves better results with less burden and faster speed.

\subsection{Ablation Study}
Ablation study on the main CSFMamba modules demonstrate the effectiveness of each computational unit. Result Table~\ref{Ablation} on the MUUFL dataset show that reducing the number of network modules will make the classification results worse. This side verifies the supplementary effect of the preprocessing module on the initial information and the effect of the CNN convolutional layer on feature extraction. Cross State Fusion Mamba plays the most critical role of cross token feature in the network, so the results are greatly reduced after removal. The results conclusively establish the ability of these key modules to assist network training and demonstrate the effectiveness of their collaboration.

\section{Discussion}
In this study, we propose a cross-state fusion training method based on the Mamba operator. This is a novel and competitive scheme to apply the new operator to object classification in remote sensing images. CSFMamba extracts the main features using reasonable information processing and uses the cross fusion between state spaces. This method has a longer range of context understanding in complex situations, and can ensure that it does not bring more burden. Experimental results on two complex HSI and LiDAR multimodal image classification datasets demonstrate the effectiveness of the proposed method. It proves that Mamba operator has a wide application prospect in remote sensing image classification. Simultaneously less burden and faster training speed could be potentially important for practical deployment of recognition networks. CSFMamba can effectively complete multimodal classification tasks, which is superior to the current existing methods in the field, marking the progress direction of the current most advanced technology.

\small
\bibliographystyle{IEEEtranN}
\bibliography{main}

\begin{thebibliography}{18}
\providecommand{\natexlab}[1]{#1}
\providecommand{\url}[1]{#1}
\csname url@samestyle\endcsname
\providecommand{\newblock}{\relax}
\providecommand{\bibinfo}[2]{#2}
\providecommand{\BIBentrySTDinterwordspacing}{\spaceskip=0pt\relax}
\providecommand{\BIBentryALTinterwordstretchfactor}{4}
\providecommand{\BIBentryALTinterwordspacing}{\spaceskip=\fontdimen2\font plus
\BIBentryALTinterwordstretchfactor\fontdimen3\font minus \fontdimen4\font\relax}
\providecommand{\BIBforeignlanguage}[2]{{%
\expandafter\ifx\csname l@#1\endcsname\relax
\typeout{** WARNING: IEEEtranN.bst: No hyphenation pattern has been}%
\typeout{** loaded for the language `#1'. Using the pattern for}%
\typeout{** the default language instead.}%
\else
\language=\csname l@#1\endcsname
\fi
#2}}
\providecommand{\BIBdecl}{\relax}
\BIBdecl

\bibitem[Yue et~al.(2022)Yue, Fang, and He]{yue2022Spectral}
J.~Yue, L.~Fang, and M.~He, ``Spectral-spatial latent reconstruction for open-set hyperspectral image classification,'' \emph{IEEE Transactions on Image Processing}, vol.~31, pp. 5227--5241, 2022.

\bibitem[Jiang et~al.(2022)Jiang, Li, Li, Xiao, and Zhou]{jiang2022afusion}
N.~Jiang, H.-B. Li, C.-J. Li, H.-X. Xiao, and J.-W. Zhou, ``A fusion method using terrestrial laser scanning and unmanned aerial vehicle photogrammetry for landslide deformation monitoring under complex terrain conditions,'' \emph{IEEE Transactions on Geoscience and Remote Sensing}, vol.~60, pp. 1--14, 2022.

\bibitem[Xu et~al.(2024)Xu, Jiang, Zhou, Li, Liu, and Lin]{xu2023robust}
G.~Xu, X.~Jiang, Y.~Zhou, S.~Li, X.~Liu, and P.~Lin, ``Robust land cover classification with multimodal knowledge distillation,'' \emph{IEEE Transactions on Geoscience and Remote Sensing}, vol.~62, pp. 1--16, 2024.

\bibitem[Xu et~al.(2023)Xu, Jiang, Li, Zhang, and Liu]{xu2023exploring}
G.~Xu, X.~Jiang, X.~Li, Z.~Zhang, and X.~Liu, ``Exploring self-supervised learning for multi-modal remote sensing pre-training via asymmetric attention fusion,'' \emph{Remote Sensing}, vol.~15, no.~24, p. 5682, 2023.

\bibitem[Melgani and Bruzzone(2004)]{melgani2004classification}
F.~Melgani and L.~Bruzzone, ``Classification of hyperspectral remote sensing images with support vector machines,'' \emph{IEEE Transactions on Geoscience and Remote Sensing}, vol.~42, no.~8, pp. 1778--1790, 2004.

\bibitem[Ahmad et~al.(2022)Ahmad, Shabbir, Roy, Hong, Wu, Yao, Khan, Mazzara, Distefano, and Chanussot]{ahmad2021hyper}
M.~Ahmad, S.~Shabbir, S.~K. Roy, D.~Hong, X.~Wu, J.~Yao, A.~M. Khan, M.~Mazzara, S.~Distefano, and J.~Chanussot, ``Hyperspectral image classification—traditional to deep models: A survey for future prospects,'' \emph{IEEE Journal of Selected Topics in Applied Earth Observations and Remote Sensing}, vol.~15, pp. 968--999, 2022.

\bibitem[Hong et~al.(2022)Hong, Gao, Hang, Zhang, and Chanussot]{hong2020deep}
D.~Hong, L.~Gao, R.~Hang, B.~Zhang, and J.~Chanussot, ``Deep encoder–decoder networks for classification of hyperspectral and lidar data,'' \emph{IEEE Geoscience and Remote Sensing Letters}, vol.~19, pp. 1--5, 2022.

\bibitem[Hang et~al.(2020)Hang, Li, Ghamisi, Hong, Xia, and Liu]{hang2020classification}
R.~Hang, Z.~Li, P.~Ghamisi, D.~Hong, G.~Xia, and Q.~Liu, ``Classification of hyperspectral and lidar data using coupled cnns,'' \emph{IEEE Transactions on Geoscience and Remote Sensing}, vol.~58, no.~7, pp. 4939--4950, 2020.

\bibitem[Zhao et~al.(2022)Zhao, Tao, Li, Philips, and Liao]{zhao2021fractional}
X.~Zhao, R.~Tao, W.~Li, W.~Philips, and W.~Liao, ``Fractional gabor convolutional network for multisource remote sensing data classification,'' \emph{IEEE Transactions on Geoscience and Remote Sensing}, vol.~60, pp. 1--18, 2022.

\bibitem[Wu and Prasad(2017)]{wu2017convolutional}
H.~Wu and S.~Prasad, ``Convolutional recurrent neural networks for hyperspectral data classification,'' \emph{Remote Sensing}, vol.~9, no.~3, p. 298, 2017.

\bibitem[Vaswani(2017)]{vaswani2017attention}
A.~Vaswani, ``Attention is all you need,'' \emph{Advances in Neural Information Processing Systems}, 2017.

\bibitem[Zhao et~al.(2023)Zhao, Ye, Sun, Wu, Pan, and Jeon]{zhao2022joint}
G.~Zhao, Q.~Ye, L.~Sun, Z.~Wu, C.~Pan, and B.~Jeon, ``Joint classification of hyperspectral and lidar data using a hierarchical cnn and transformer,'' \emph{IEEE Transactions on Geoscience and Remote Sensing}, vol.~61, pp. 1--16, 2023.

\bibitem[Li et~al.(2023)Li, Wang, Wang, Liu, Wu, and Wang]{li2023mixing}
K.~Li, D.~Wang, X.~Wang, G.~Liu, Z.~Wu, and Q.~Wang, ``Mixing self-attention and convolution: A unified framework for multisource remote sensing data classification,'' \emph{IEEE Transactions on Geoscience and Remote Sensing}, vol.~61, pp. 1--16, 2023.

\bibitem[Gu and Dao(2023)]{gu2023mamba}
A.~Gu and T.~Dao, ``Mamba: Linear-time sequence modeling with selective state spaces,'' \emph{arXiv preprint arXiv:2312.00752}, 2023.

\bibitem[Gu et~al.(2021)Gu, Goel, and R{\'e}]{gu2021eff}
A.~Gu, K.~Goel, and C.~R{\'e}, ``Efficiently modeling long sequences with structured state spaces,'' \emph{arXiv preprint arXiv:2111.00396}, 2021.

\bibitem[Yang et~al.(2024)Yang, Zhou, Wang, Tian, and Liew]{yang2024hsimamba}
J.~X. Yang, J.~Zhou, J.~Wang, H.~Tian, and A.~W.~C. Liew, ``Hsimamba: Hyperpsectral imaging efficient feature learning with bidirectional state space for classification,'' \emph{arXiv preprint arXiv:2404.00272}, 2024.

\bibitem[Huang et~al.(2024)Huang, Chen, and He]{huang2024spectral}
L.~Huang, Y.~Chen, and X.~He, ``Spectral-spatial mamba for hyperspectral image classification,'' \emph{arXiv preprint arXiv:2404.18401}, 2024.

\bibitem[Gao et~al.(2024)Gao, Jin, Zhou, Dong, and Du]{gao2024multi}
F.~Gao, X.~Jin, X.~Zhou, J.~Dong, and Q.~Du, ``Msfmamba: Multi-scale feature fusion state space model for multi-source remote sensing image classification,'' \emph{arXiv preprint arXiv:2408.14255}, 2024.

\end{thebibliography}

\end{document}